\documentclass[10pt,twocolumn,letterpaper]{article}

\usepackage[pagenumbers]{cvpr} 

\definecolor{cvprblue}{rgb}{0.21,0.49,0.74}
\usepackage[pagebackref,breaklinks,colorlinks,allcolors=cvprblue]{hyperref}
\usepackage{multirow}
\usepackage{bbding}
\usepackage{array} 
\newcolumntype{L}[1]{>{\raggedright\arraybackslash}m{#1}}
\newcolumntype{C}[1]{>{\centering\arraybackslash}m{#1}}
\usepackage{makecell}
\usepackage{amsmath}
\usepackage{bm}
\usepackage{xcolor}

\def\paperID{46417} 
\def\confName{CVPR}
\def\confYear{2026}

\title{Compressed-Domain-Aware Online Video Super-Resolution}

\author{
Yuhang Wang$^{1}$, Hai Li$^{1,2}$, Shujuan Hou$^{1,2,}$\thanks{Corresponding author} , Zhetao Dong$^{1}$, Xiaoyao Yang$^{1}$\\
$^{1}$School of Information and Electronics, Beijing Institute of Technology, Beijing, China\\
$^{2}$Terahertz Science and Application Center, Beijing Institute of Technology, Zhuhai, Guangdong, China\\
{\tt\small \{yuhangwang, haili, shujuanhou, 3120230750, xyyang\}@bit.edu.cn}
}

\begin{document}
\maketitle
\begin{abstract}
In bandwidth-limited online video streaming, videos are usually downsampled and compressed. Although recent online video super-resolution (online VSR) approaches achieve promising results, they are still compute-intensive and fall short of real-time processing at higher resolutions, due to complex motion estimation for alignment and redundant processing of consecutive frames. To address these issues, we propose a compressed-domain-aware network (CDA-VSR) for online VSR, which utilizes compressed-domain information, including motion vectors, residual maps, and frame types to balance quality and efficiency. Specifically, we propose a motion-vector-guided deformable alignment module that uses motion vectors for coarse warping and learns only local residual offsets for fine-tuned adjustments, thereby maintaining accuracy while reducing computation. Then, we utilize a residual map gated fusion module to derive spatial weights from residual maps, suppressing mismatched regions and emphasizing reliable details. Further, we design a frame-type-aware reconstruction module for adaptive compute allocation across frame types, balancing accuracy and efficiency. On the REDS4 dataset, our CDA-VSR surpasses the state-of-the-art method TMP, with a maximum PSNR improvement of \textbf{0.13 dB} while delivering more than \textbf{double} the inference speed. The code will be released at \url{https://github.com/sspBIT/CDA-VSR}.
\end{abstract}

\section{Introduction}
\label{sec:intro}
Video super-resolution (VSR) aims to reconstruct the high-resolution (HR) video sequence from low-resolution (LR) frames. With the rise of online applications such as video conferencing and live streaming, online VSR has garnered increasing attention \cite{fuoli2023fast, xiao2023online, zhang2024tmp, yin2024online}. In online VSR, the current frame is enhanced using only past and current frames under a strict time budget.

Recently, a series of online VSR methods have been proposed \cite{tang2025multi, menon2024video, zhu2025trajectory, xiao2023online, sajjadi2018frame, isobe2020video, isobe2020revisiting}. Flow-based alignment methods \cite{sajjadi2018frame, xia2023structured} improve super-resolution (SR) quality by accurately aligning frames using optical flow, but optical flow estimation is computationally intensive. Implicit alignment methods \cite{isobe2020video, isobe2020revisiting, fuoli2019efficient} improve efficiency at the expense of reconstruction quality, most notably under large motions.

To better balance accuracy and efficiency, recent works explore efficient alignment and fusion modules \cite{zhang2024tmp, fuoli2023fast, tang2025multi, xiao2023online, jin2023kernel}. DAP \cite{fuoli2023fast} uses a deformable attention pyramid to efficiently align features from the previous frame with those of the current frame. TMP \cite{zhang2024tmp} leverages the motion continuity between frames, propagates the offsets across frames and refines them locally, thus avoiding redundant computations. However, these methods still struggle with complex motion estimation and redundant processing of consecutive frames, leading to a heavy computational burden, particularly at higher resolutions such as 2K.

\begin{figure}[t]
	\centering
	\includegraphics[width=3.2 in]{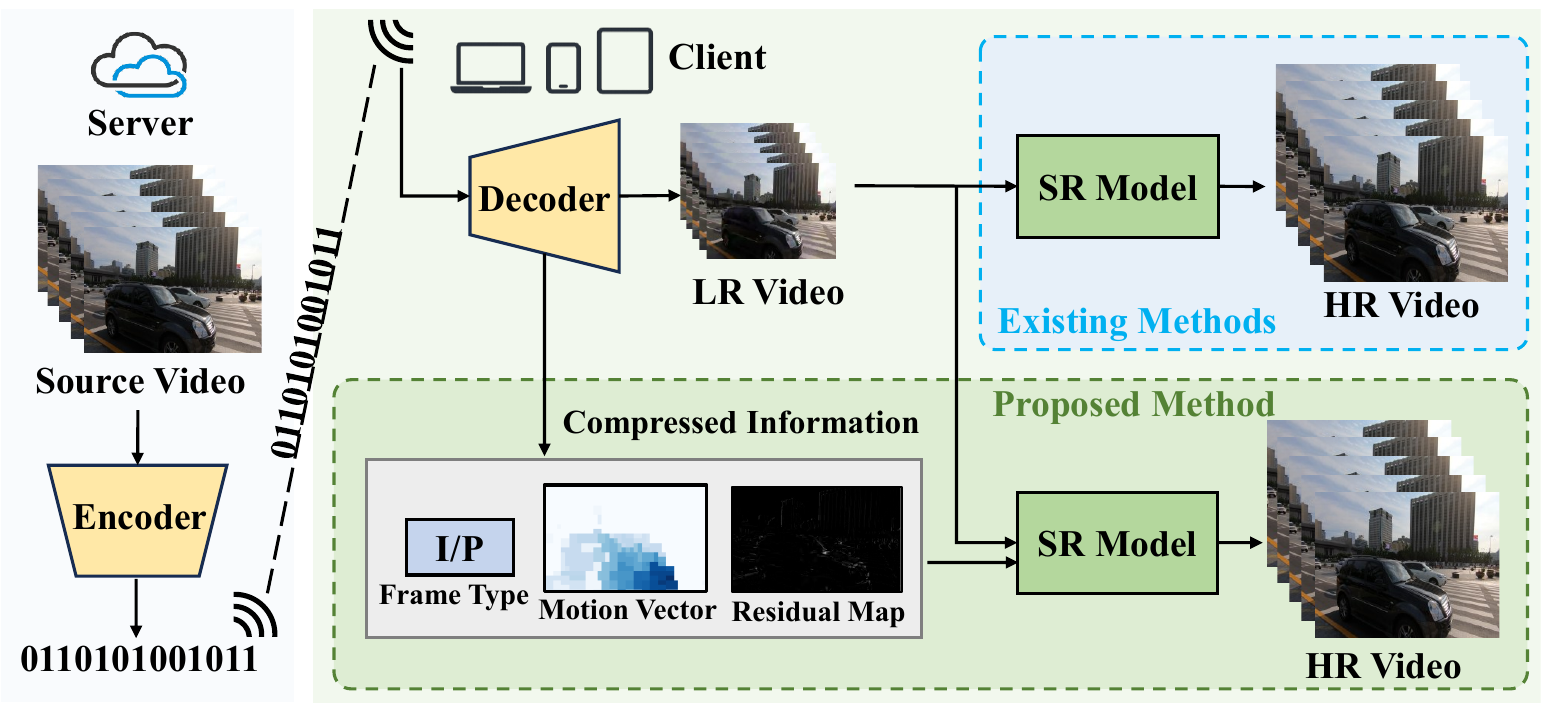}
	\caption{Comparison between existing online VSR methods and our proposed method. The server downsamples and encodes the source video, and then transmits the compressed stream. The client decodes and performs super-resolution. Our method uses compressed-domain information, including frame type, motion vectors, and residual maps, to guide alignment, fusion, and reconstruction, improving both accuracy and efficiency.} 
	\label{fig1}
\end{figure}

\begin{figure}[t]
	\centering
	\includegraphics[width=3 in]{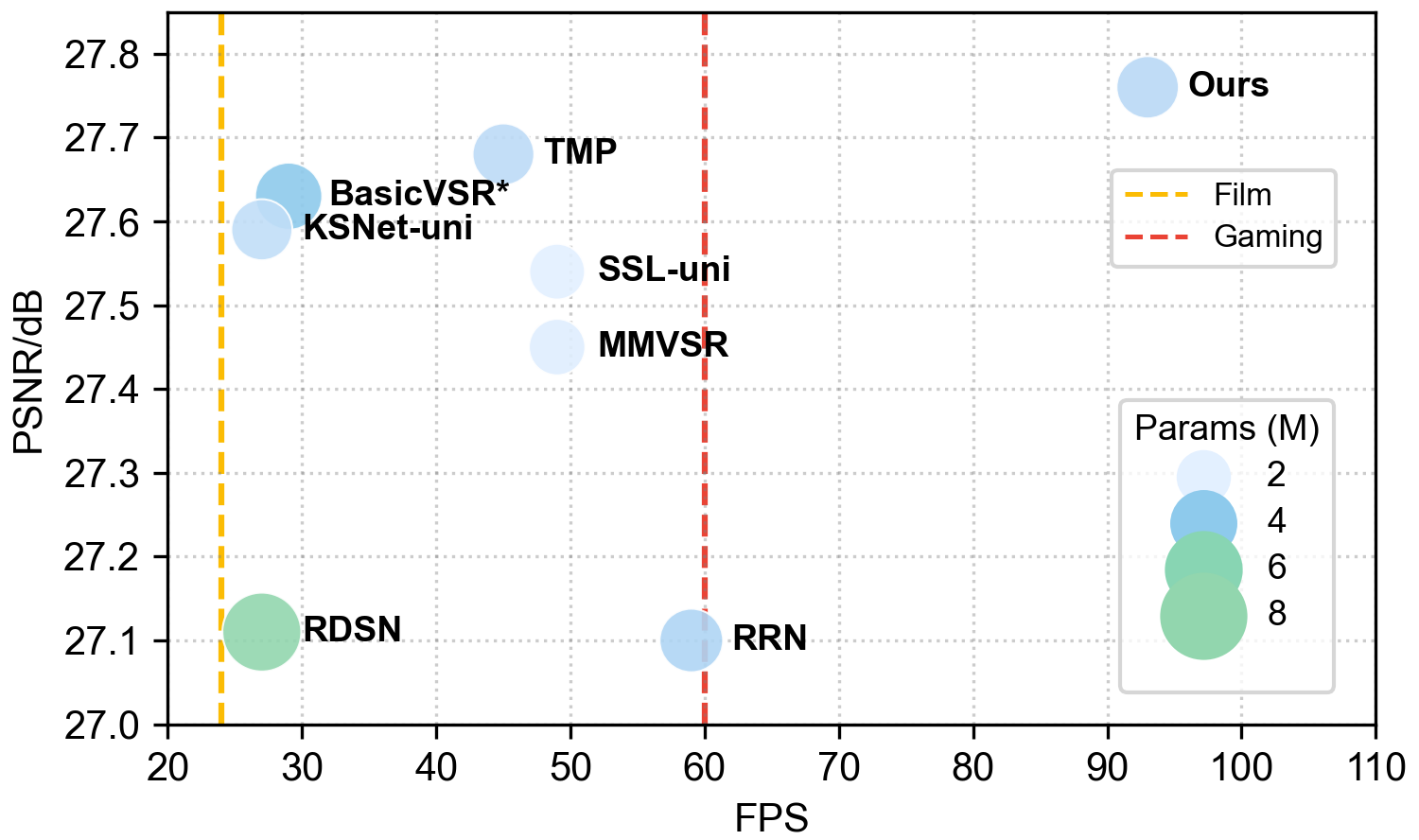}
	\caption{PSNR, FPS, and Parameters of different methods on REDS4 for 4$\times$ upscaling at CRF=18.} 
	\label{fig2}
\end{figure}

These methods typically rely solely on the LR video frames and do not exploit the valuable compressed-domain information such as motion vectors, residual maps, and frame types, readily available in the bitstream. Motion vectors describe coarse inter-frame motion, residual maps reflect high-frequency differences, and frame types determine inter-frame reference relationships. Leveraging this extra information can further improve both accuracy and efficiency in online VSR.

Building upon this idea, we propose a compressed-domain-aware online VSR framework (CDA-VSR), as illustrated in Figure \ref{fig1}. Our CDA-VSR is designed with three key modules that exploit the distinct characteristics of motion vectors, residual maps, and frame types. 
We propose a motion-vector-guided deformable alignment module (MVGDA) that uses motion vectors for coarse alignment, then initializes deformable offsets for local refinement. This design addresses the limitations of flow-based alignment methods (which are computationally expensive) and implicit alignment methods (which struggle with large motion). Then, instead of concatenating inter-frame features \cite{zhang2024tmp, wu2024real, viswanathan2025low}, which propagates mismatched details from the previous frame, we propose residual map gated fusion (RMGF). The residual map predicts spatial weights that suppress irrelevant regions and emphasize reliable structures to improve reconstruction quality. Moreover, we introduce frame-type-aware reconstruction (FTAR): a high-capacity path for I-frames and a lightweight path for P-frames. This frame-type adaptive allocation preserves keyframe fidelity, avoids redundant computation on P-frames, and significantly enhances real-time processing efficiency. 

The main contributions are summarized as follows.
\begin{itemize}
  \item We propose a compressed-domain-aware online VSR framework (CDA-VSR). By leveraging motion vectors, residual maps, and frame types to guide frame alignment, fusion, and reconstruction, CDA-VSR improves both accuracy and efficiency.
  \item We design a motion-vector-guided deformable alignment module (MVGDA) and a residual map gated fusion module (RMGF). MVGDA combines coarse motion-vector alignment with local deformable refinement, maintaining pixel-level accuracy with reduced complexity. RMGF uses residual maps to generate spatial weights, suppressing misaligned regions and enhancing detail reliability.
  \item We propose a frame-type-aware reconstruction strategy (FTAR). I-frames are processed with a high-capacity reconstruction module to preserve global fidelity, while lightweight modules are designed for P-frames to accelerate inference.
  \item Experiments on the REDS4 dataset show that our method achieves approximately $90$ FPS while maintaining visual quality comparable to state-of-the-art methods and delivering $>2\times$ the inference speed, as illustrated in Figure~\ref{fig2}.
\end{itemize}

\section{Related Works}
\noindent \textbf{Video Super-Resolution.}
VSR aims to reconstruct a HR video sequence from degraded LR inputs. Based on the paradigms, VSR methods \cite{zhou2024upscale, yi2019progressive, wang2019edvr, xu2025videogigagan, shang2024arbitrary, li2024savsr, xie2025star, du2025patchvsr,liang2024vrt} can be roughly grouped into sliding-window based VSR and recurrent based VSR. Sliding-window based VSR \cite{wang2019edvr, caballero2017real, kappeler2016video, xue2019video, wang2020deep, yi2019progressive, tian2020tdan, li2020mucan} uses a fixed set of neighboring frames to reconstruct one or more target frames. The available information is constrained by the window size, so these methods can only exploit temporal details within a restricted subset of the video. To exploit a longer temporal context, recurrent based VSR methods \cite{chan2021basicvsr, chan2022basicvsr++, shi2022rethinking, liang2022recurrent, zhang2024realviformer, sajjadi2018frame, isobe2020video,isobe2020revisiting, fuoli2019efficient} reuse information by propagating hidden states and reconstructed frames across time. Bidirectional propagation methods \cite{chan2021basicvsr, chan2022basicvsr++, zhou2024video, shi2025self} further boost accuracy by passing information from both past and future frames. By leveraging many support frames, these methods typically achieve higher accuracy at the cost of increased latency. Unidirectional propagation methods \cite{sajjadi2018frame, fuoli2019efficient, isobe2020video, isobe2020revisiting, liang2022recurrent} aggregate the information from the past and current frames, as well as several cached future frames, thereby improving efficiency. When only past and current frames are available, unidirectional propagation methods are suitable for online VSR.

\begin{figure*}[t!]
	\centering
	\includegraphics[width=7 in]{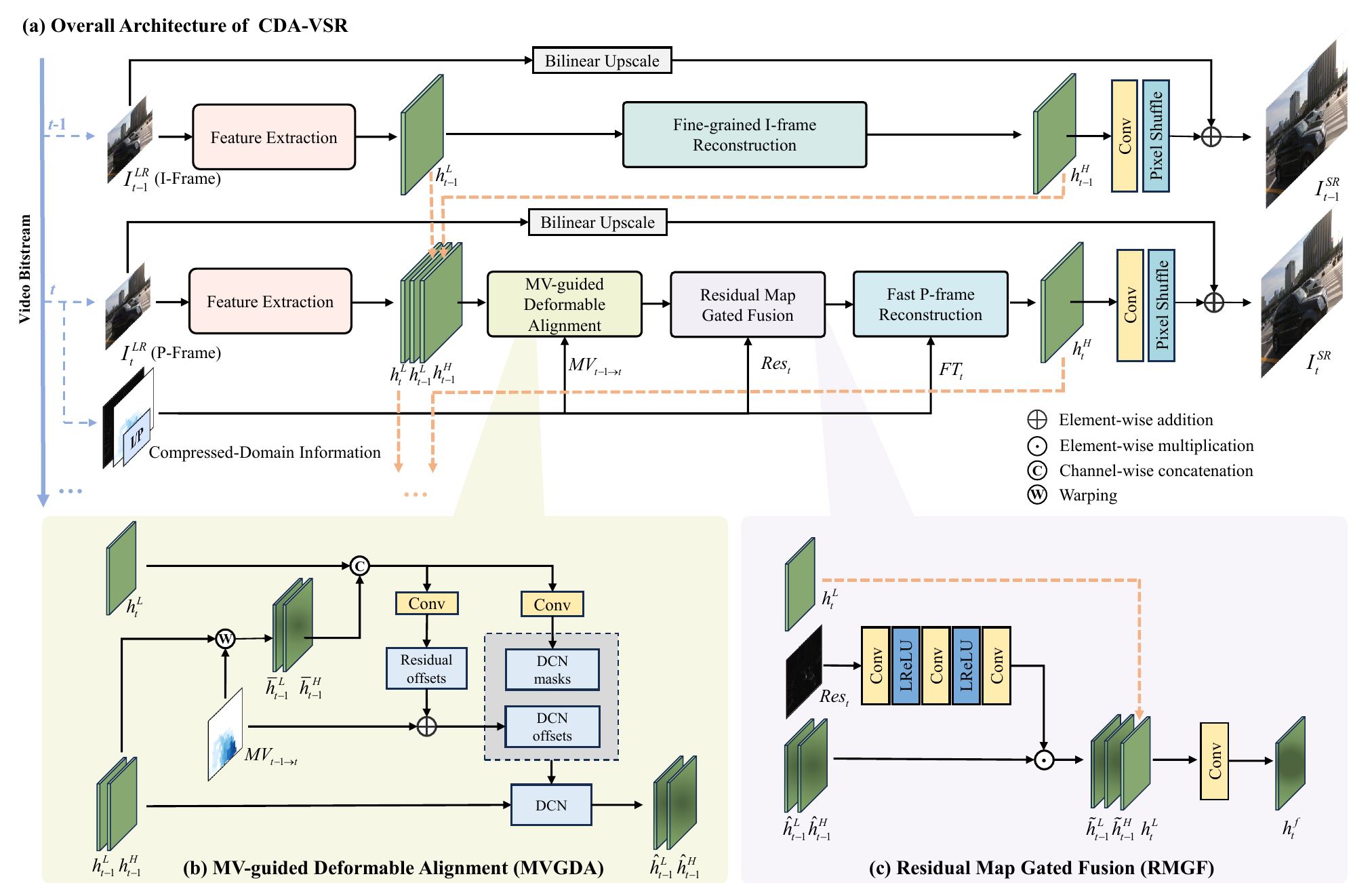}
	\caption{Overall architecture of the proposed Compressed-Domain-Aware VSR framework (CDA-VSR). 
Given LR frames and compressed-domain information (motion vectors, residual maps, and frame types), CDA-VSR reconstructs the corresponding HR frames through three key modules: 
(1) the MV-guided Deformable Alignment (MVGDA); (2) the Residual Map Gated Fusion (RMGF);
(3) the Frame-Type-Aware Reconstruction (FTAR) with two branches, Fine-grained I-Frame Reconstruction and Fast P-Frame Reconstruction.} 
	\label{fig3}
\end{figure*}

\noindent \textbf{Online Video Super-Resolution.}
Online VSR requires real-time reconstruction of the current frame during video playback \cite{xiao2023online}. This imposes causal constraints (only past and current frames can be used) and demands low latency, in contrast to offline VSR, which can exploit bidirectional information. Early online strategies apply lightweight single-frame SR methods \cite{dong2015image, wang2023omni,zheng2024smfanet,liang2021swinir,dong2025lightweight,liu2025catanet}. By processing each frame independently, they fail to exploit temporal redundancy, leading to limited improvements in reconstruction quality. Later works explore unidirectional recurrent architectures that reuse information from previous frames, such as RSDN \cite{isobe2020video} and RRN \cite{isobe2020revisiting}. Recent works \cite{zhang2024tmp, fuoli2023fast, tang2025multi, xiao2023online, jin2023kernel, zhu2025trajectory} improve online VSR mainly through enhanced alignment and fusion modules. KSNet \cite{jin2023kernel} uses a multi-flow deformable alignment module and a kernel-split strategy. TMP \cite{zhang2024tmp} exploits inter-frame motion similarity, propagates estimated motion fields across frames, and incrementally refines them, thereby avoiding redundant motion estimation for each frame. MMVSR \cite{tang2025multi} performs dynamic-static decoupled alignment and fuses multi-memory streams to improve long-range temporal modeling. However, existing approaches still incur substantial compute and latency due to complex motion estimation for alignment and redundant processing across consecutive frames, which are amplified at higher resolutions (e.g., 2K).

\noindent \textbf{Compressed-Domain Information for VSR.}
Recently, many vision tasks \cite{chen2021fast,wang2019fast,liu2024vadiffusion,fang2023you} have benefited from compressed-domain information. Similarly, a few methods have attempted to incorporate such information into VSR \cite{chen2021compressed, zhang2022codec, wang2023compression}. CDVSR \cite{chen2021compressed} fuses bitstream coding priors with deep SR models to better restore textures and details. CIAF \cite{zhang2022codec} uses motion vectors as optical-flow proxies and residual maps to identify static regions, thereby skipping redundant processing and improving efficiency. CAVSR \cite{wang2023compression} employs a compression encoder to extract compression-level features and a modulation module to adapt SR to varying compression strengths. These methods show that incorporating compressed-domain information (e.g., motion vectors, residual maps) can improve VSR performance. However, as they are not designed for online VSR, their inference speed remains insufficient for real-time applications. In addition, existing methods have not fully explored dedicated designs for different types of compressed-domain information, which limits the potential gains achievable from such information. In contrast, CDA-VSR tailors specialized modules to the characteristics of each type of compressed-domain information, while explicitly satisfying the causality and real-time constraints of online VSR.

\section{Methodology}
\subsection{Overall Architecture}
Online VSR aims to reconstruct the HR reference frame $I_t^{HR} \in \mathbb{R}^{sH \times sW \times 3}$ from its corresponding LR frame $I_t^{LR} \in \mathbb{R}^{H \times W \times 3}$ 
and the $N$ supporting frames $I^{LR}_{[t-N:t-1]}$, where $s$ denotes the upsampling factor, 
$H$ and $W$ denote the height and width of LR frames, and $t$ represents the timestamp of the video stream. 
Unlike existing online VSR methods that only exploit decoded LR frames, our CDA-VSR additionally leverages compressed-domain priors from the bitstream, including motion vectors $MV_{t-1\rightarrow t}$, residual maps $Res_t$, and frame types $FT_t$, as shown in Fig.~\ref{fig3}. These priors provide motion cues for feature alignment, suppress misaligned regions for more reliable fusion, and distinguish frame types for adaptive reconstruction.

Building on earlier works \cite{fuoli2023fast, zhang2024tmp}, CDA-VSR adopts a recurrent structure, which enhances computational efficiency and meets the requirements of real-time processing. Following existing works \cite{fuoli2023fast,zhang2024tmp}, we adopt a shallow feature extraction network to map each decoded LR frame into latent features $h_t^{L}$. CDA-VSR then consists of three key modules designed to leverage the characteristics of different compressed-domain information. First, the MV-guided deformable alignment module (MVGDA) employs motion vectors for coarse alignment, followed by a lightweight deformable convolution to refine local misalignments. Second, the residual map gated fusion module (RMGF) generates spatially varying weights from residual maps to suppress irrelevant details and enhance reliable regions. Third, the frame-type-aware reconstruction (FTAR) adaptively allocates computational resources. I-frames are reconstructed using a fine-grained branch to preserve global fidelity, while P-frames are processed by a lightweight branch to improve inference speed. These three modules are described in detail in Sections~\ref{Alignment}, \ref{Fusion}, and \ref{Reconstruction}.

In summary, CDA-VSR exploits the complementary roles of motion vectors, residual maps, and frame types to achieve a superior trade-off between reconstruction quality and computational efficiency, enabling real-time online VSR.

\subsection{MV-guided Deformable Alignment (MVGDA)}
\label{Alignment}
Accurate and efficient frame alignment is crucial in online VSR to exploit temporal redundancy and improve reconstruction quality. Flow-based alignment explicitly estimates optical flow to warp frames or features, but is computationally expensive and sensitive to flow errors, whereas deformable convolutions implicitly compensate motion via learned sampling offsets but still struggle with large or complex displacements due to unconstrained offsets.

To address these limitations, we leverage motion vectors (MVs) extracted directly from the video bitstream as temporal priors. MVs describe block-level displacements between adjacent frames and can be obtained essentially for free during decoding, providing a computationally efficient alternative to optical flow. Let $h_{t-1}$ and $h_t$ denote the features of the previous and current frames, respectively. 
We first perform coarse alignment by warping the previous-frame features $h_{t-1}$ using MVs:
\begin{equation}
\overline{h}_{t-1} = \mathcal{W}(h_{t-1}; MV_{t-1 \rightarrow t}),
\end{equation}
where $\mathcal{W}(\cdot)$ denotes the warping operator. 
This step efficiently compensates large inter-frame motion.

Although MVs provide useful motion priors, their block-wise nature forces all pixels within a block to share a single motion vector, ignoring intra-block motion variations. As a result, MVs become unreliable near object boundaries and in regions with complex or non-rigid motion.
To mitigate this, we embed MVs into a deformable convolution (DCN). 
Specifically, the DCN offsets are initialized with motion vectors $o_{MV}$, and a lightweight convolutional network predicts residual offsets $\Delta o$ to further refine them. 
In parallel, a modulation mask $m$ is predicted to adaptively re-weight the sampling locations:
\begin{equation}
\Delta o = \mathcal{C}^{o}( [h_t, \overline{h}_{t-1}]),
\end{equation}
\begin{equation}
m = \sigma(\mathcal{C}^{m}( [h_t, \overline{h}_{t-1}] )),
\end{equation}
where $\mathcal{C}^{o}$ and $\mathcal{C}^{m}$ denote convolutional subnetworks, $[\cdot,\cdot]$ represents channel-wise concatenation, and $\sigma(\cdot)$ is the sigmoid function. The aligned features $\hat{h}_{t-1}$ are then obtained by applying DCN to the previous unwarped features $h_{t-1}$:
\begin{equation}
\hat{h}_{t-1} = \mathcal{D}\big(h_{t-1}; o_{MV} + \Delta o, m\big),
\end{equation}
where $\mathcal{D}$ denotes DCN. In this way, MVGDA uses motion vectors for efficient coarse alignment, while the deformable convolution only needs to learn local residual offsets, which simplifies offset learning and leads to more robust and efficient alignment under large and complex motions.

So far, we have described alignment for a single feature from the previous frame. In practice, our CDA-VSR exploits two complementary feature representations: coarse features $h_{t-1}^{L}$ from the encoder and fine-grained features $h_{t-1}^{H}$ from the reconstruction module. The former provide robust structural priors that are less affected by reconstruction noise, while the latter carry rich texture details that enhance fidelity. Both types of features are aligned using MVGDA.

\begin{gather}
\overline{h}_{t-1}^{L}, \overline{h}_{t-1}^{H} 
= \mathcal{W}(h_{t-1}^{L}, h_{t-1}^{H}; MV_{t-1 \rightarrow t}), \\
\Delta o = \mathcal{C}^{o}\big([h_t^{L}, \overline{h}_{t-1}^{L}, \overline{h}_{t-1}^{H}]\big), \\
m = \sigma(\mathcal{C}^{m}\big([h_t^{L}, \overline{h}_{t-1}^{L}, \overline{h}_{t-1}^{H}]\big)), \\
\hat{h}_{t-1}^{L}, \hat{h}_{t-1}^{H} 
= \mathcal{D}\big(h_{t-1}^{L}, h_{t-1}^{H}; o_{MV} + \Delta o, m\big).
\end{gather}
The two aligned features $\hat{h}_{t-1}^{L}$ and $\hat{h}_{t-1}^{H}$ are then jointly propagated to the current step, enabling the network to benefit from both stable global structures and detailed textures.

\subsection{Residual Map Gated Fusion (RMGF)}
\label{Fusion}
After aligning features across frames with MVGDA, the next step is to effectively fuse information from the previous and current frames. A simple strategy is to concatenate the aligned features along the channel dimension \cite{zhang2024tmp, wu2024real, viswanathan2025low}. However, misaligned or inconsistent regions from the previous frame can propagate errors and degrade reconstruction quality. Therefore, it is crucial to selectively exploit reliable regions from the previous frame while suppressing misleading ones.

Residual maps $Res_t$ obtained from the video bitstream represent the pixel-wise difference between the current frame and its motion-compensated prediction from reference frames. Large residual values indicate regions where motion compensation fails, such as occlusion boundaries or regions with complex local motion. Therefore, $Res_t$ naturally highlights temporally inconsistent regions and provides a useful cue for guiding feature fusion.

To leverage this cue, we design a residual map gated fusion module (RMGF). 
A lightweight network $\mathcal{F}_{\text{res}}(\cdot)$ first transforms $Res_t$ into a spatial gating map:
\begin{equation}
M_t = \sigma(\mathcal{F}_{\text{res}}(Res_t)),
\end{equation}
where $\sigma(\cdot)$ denotes the sigmoid function.
The gate $M_t$ suppresses unreliable regions in the aligned features $\hat{h}_{t-1}^{L}, \hat{h}_{t-1}^{H}$, and the fused feature $h_t^{f}$ is obtained as:
\begin{equation}
h_t^{f} = \mathcal{C}^{f}([M_t \odot \hat{h}_{t-1}^{L},\, M_t \odot \hat{h}_{t-1}^{H},\, h_t^{L}]),
\end{equation}
where $\mathcal{C}^{f}$ is a $3 \times 3$ convolution and $\odot$ denotes element-wise multiplication. 
In this way, the current-frame features $h_t^{L}$ serve as a stable baseline, while temporal features from previous frames contribute only in regions where they are reliable. By preserving trustworthy temporal information and down-weighting misaligned regions, RMGF improves the overall reconstruction quality.

\begin{table*}
\centering
\caption{Comparison with state-of-the-art online VSR methods on the REDS4 dataset. Runtime is measured on a single NVIDIA RTX 3090 with LR inputs of 320$\times$180. Following practical usage, methods with FPS $\geq$24 are regarded as film real-time, and FPS $\geq$60 as gaming real-time. The best and the second best results are colored with \textcolor{red}{red} and \textcolor{blue}{blue}.}
\label{tab1}
\resizebox{\linewidth}{!}{
\begin{tabular}{L{2.6cm}C{1.3cm}C{1.3cm}C{1.3cm}C{0.9cm}C{1.1cm}C{0.9cm}C{1.0cm}C{1.0cm}C{1.0cm}C{1.0cm}C{1.0cm}C{1.0cm}C{1.0cm}}
\toprule
\multirow{2}{*}{Method} & \multirow{2}{*}{\shortstack{Film R.T.\\ (FPS$\geq$24)}} & \multirow{2}{*}{\shortstack{Game R.T.\\ (FPS$\geq$60)}} & \multirow{2}{*}{\shortstack{Runtime $\downarrow$ \\ (ms)}} & \multirow{2}{*}{\shortstack{FPS $\uparrow$\\(1/s)}} & \multirow{2}{*}{\shortstack{Params $\downarrow$ \\ (M)}} & \multirow{2}{*}{\shortstack{MACs $\downarrow$ \\(G)}} 
& \multicolumn{5}{c}{CRF18 (PSNR(dB)$\uparrow$ / SSIM$\uparrow$ / LPIPS$\downarrow$)}
& \multicolumn{1}{c}{CRF23}
& \multicolumn{1}{c}{CRF28} \\
\cmidrule(lr){8-12}\cmidrule(lr){13-13}\cmidrule(lr){14-14}
& & & & & & & clip000 & clip011 & clip015 & clip020 &Avg &Avg &Avg \\
\midrule
BasicVSR$^{*}$ \cite{chan2021basicvsr}& \Checkmark & \XSolidBrush & 34.5 & 29 & 4.0 & 254
& \shortstack{\textcolor{blue}{25.72}\\\textcolor{blue}{0.7040}\\0.3435} & \shortstack{27.91\\0.7739\\0.3371} & \shortstack{30.26\\0.8432\\0.3239} & \shortstack{26.62\\0.7730\\0.3343} & \shortstack{27.63\\0.7735\\0.3347}
& \shortstack{26.54\\0.7326\\0.3809} & \shortstack{25.13\\0.6795\\0.4335} \\
\midrule
RRN \cite{isobe2020revisiting} & \Checkmark & \XSolidBrush & \textcolor{blue}{16.9} & \textcolor{blue}{59} & 3.4 & 193
& \shortstack{25.21\\0.6748\\0.3645} & \shortstack{27.32\\0.7546\\0.3518} & \shortstack{29.76\\0.8223\\0.3311} & \shortstack{26.11\\0.7553\\0.3480} & \shortstack{27.10\\0.7518\\0.3489}
& \shortstack{26.22\\0.7192\\0.3915} & \shortstack{24.96\\0.6718\\0.4410} \\
\midrule
RSDN \cite{isobe2020video} & \Checkmark & \XSolidBrush & 37.0 & 27 & 6.2 & 356
& \shortstack{25.27\\0.6782\\0.3634} & \shortstack{27.31\\0.7546\\0.3534} & \shortstack{29.76\\0.8325\\0.3305} & \shortstack{26.11\\0.7550\\0.3492} & \shortstack{27.11\\0.7551\\0.3491}
& \shortstack{26.22\\0.7198\\0.3922} & \shortstack{24.96\\0.6721\\0.4414} \\
\midrule
SSL-uni \cite{xia2023structured} & \Checkmark & \XSolidBrush & 20.4 & 49 & \textcolor{red}{2.2} & \textcolor{blue}{92}
& \shortstack{25.64\\0.7013\\0.3489} & \shortstack{27.84\\0.7712\\0.3430} & \shortstack{30.17\\0.8420\\0.3281} & \shortstack{26.49\\0.7684\\0.3434} & \shortstack{27.54\\0.7707\\0.3409}
& \shortstack{26.48\\0.7302\\0.3872} & \shortstack{25.09\\0.6781\\0.4395} \\
\midrule
KSNet-uni \cite{jin2023kernel} & \Checkmark & \XSolidBrush & 29.4 & 34 & 3.0 & 148
& \shortstack{25.71\\0.7035\\0.3376} & \shortstack{27.97\\0.7724\\0.3304} & \shortstack{29.93\\0.8302\\0.3312} & \shortstack{26.72\\0.7717\\0.3307} & \shortstack{27.58\\0.7695\\0.3325}
& \shortstack{26.57\\0.7327\\0.3824} & \shortstack{25.12\\0.6749\\0.4327} \\
\midrule
MMVSR \cite{tang2025multi} & \Checkmark & \XSolidBrush & 23.2 & 43 & \textcolor{blue}{2.3} & 122
& \shortstack{25.55\\0.6911\\0.3677} & \shortstack{27.72\\0.7666\\0.3498} & \shortstack{30.08\\0.8396\\0.3346} & \shortstack{26.44\\0.7648\\0.3501} & \shortstack{27.45\\0.7655\\0.3506}
& \shortstack{26.42\\0.7257\\0.3961} & \shortstack{25.04\\0.6740\\0.4478} \\
\midrule
TMP \cite{zhang2024tmp} & \Checkmark & \XSolidBrush & 22.2 & 45 & 3.1 & 176
& \shortstack{25.70\\0.7036\\\textcolor{blue}{0.3371}} & \shortstack{\textcolor{blue}{27.99}\\\textcolor{blue}{0.7750}\\\textcolor{blue}{0.3303}} & \shortstack{\textcolor{blue}{30.31}\\\textcolor{blue}{0.8437}\\\textcolor{blue}{0.3181}} & \shortstack{\textcolor{blue}{26.73}\\\textcolor{blue}{0.7764}\\\textcolor{blue}{0.3235}} & \shortstack{\textcolor{blue}{27.68}\\\textcolor{blue}{0.7747}\\\textcolor{blue}{0.3273}}
& \shortstack{\textcolor{blue}{26.58}\\\textcolor{blue}{0.7336}\\\textcolor{blue}{0.3748}} & \shortstack{\textcolor{blue}{25.17}\\\textcolor{blue}{0.6805}\\\textcolor{blue}{0.4288}} \\
\midrule
CDA-VSR (Ours) & \Checkmark & \Checkmark & \textcolor{red}{10.8} & \textcolor{red}{93} & 3.3 & \textcolor{red}{78}
& \shortstack{\textcolor{red}{25.81}\\\textcolor{red}{0.7085}\\\textcolor{red}{0.3324}} & \shortstack{\textcolor{red}{28.11}\\\textcolor{red}{0.7788}\\\textcolor{red}{0.3285}} & \shortstack{\textcolor{red}{30.32}\\\textcolor{red}{0.8455}\\\textcolor{red}{0.3171}} & \shortstack{\textcolor{red}{26.79}\\\textcolor{red}{0.7788}\\\textcolor{red}{0.3223}} & \shortstack{\textcolor{red}{27.76}\\\textcolor{red}{0.7779}\\\textcolor{red}{0.3251}}
& \shortstack{\textcolor{red}{26.70}\\\textcolor{red}{0.7384}\\\textcolor{red}{0.3705}} & \shortstack{\textcolor{red}{25.30}\\\textcolor{red}{0.6869}\\\textcolor{red}{0.4230}} \\
\bottomrule
\end{tabular}
}
\end{table*}

\begin{figure*}[t!]
	\centering
	\includegraphics[width=6.85 in]{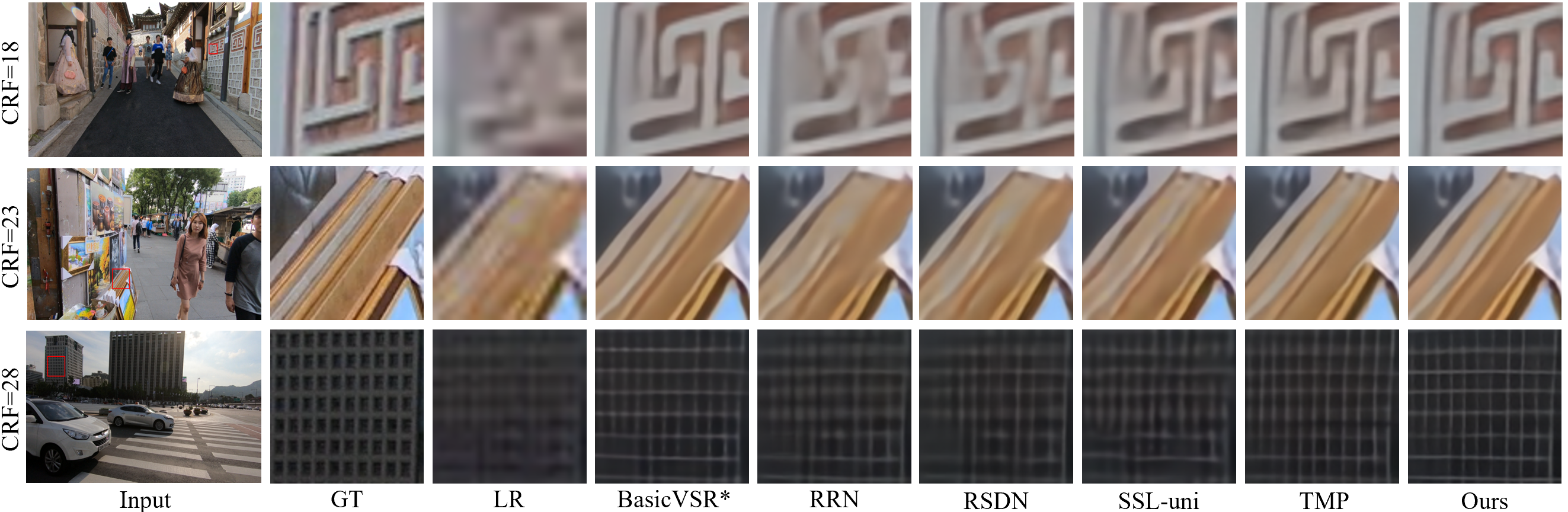}
	\caption{Qualitative comparison of different online VSR methods on the REDS4 dataset.} 
	\label{fig4}
\end{figure*}

\subsection{Frame-Type-Aware Reconstruction (FTAR)}
\label{Reconstruction}

Videos are typically encoded as a mixture of intra-coded frames (I-frames) and predictive frames (P-frames). As we target online streaming, B-frames (requiring future frames) are not considered.
I-frames contain full spatial information and serve as key references for subsequent decoding, whereas P-frames mainly store incremental updates with respect to previously decoded frames and occur much more frequently.
In online VSR, reconstructing all frames with the same computational budget is inefficient: allocating excessive computation to P-frames leads to redundant cost, while insufficient modeling of I-frames may degrade the quality of the entire sequence. 

To balance accuracy and efficiency, we propose a frame-type-aware reconstruction (FTAR) strategy that allocates computation according to the frame type. 
For I-frames, we employ a fine-grained reconstruction branch $\mathcal{R}_I$ with higher capacity to preserve global structures and maximize visual fidelity:
\begin{equation}
I_t^{SR} = \mathcal{R}_{I} (h_t^{L}), \quad \text{if } FT_t = \text{I},
\end{equation}
where $h_t^{L}$ denotes the features of the current LR frame. 
For P-frames, we adopt a fast reconstruction branch $\mathcal{R}_P$ to accelerate inference while maintaining sufficient detail:
\begin{equation}
I_t^{SR} = \mathcal{R}_{P} (h_t^{f}), \quad \text{if } FT_t = \text{P},
\end{equation}
where $h_t^{f}$ is the fused feature obtained from RMGF. 

Concretely, we use the depth of residual blocks as a proxy for computational complexity and instantiate a fine-grained branch with $m$ blocks for I-frames and a lightweight branch with $n$ $(n<m)$ blocks for P-frames. This adaptive reconstruction strategy ensures that I-frames deliver high-quality details that benefit subsequent temporal propagation, while avoiding redundant computation on the more frequent P-frames.

\subsection{Loss Function}
\label{loss}

Similar to existing works \cite{zhang2024tmp, xiao2023online, zhu2025trajectory}, we adopt the widely used Charbonnier loss~\cite{lai2017deep}.
Given the predicted high-resolution frame $I_t^{SR}$ and the corresponding ground-truth frame $I_t^{GT}$, 
the Charbonnier loss is defined as:
\begin{equation}
\mathcal{L} =
\frac{1}{T} \sum_{t=1}^{T} \sqrt{\left(I_t^{SR} - I_t^{GT}\right)^2 + \epsilon^2},
\end{equation}
where $T$ is the number of input LR frames and $\epsilon$ is a hyperparameter.

\section{Experiments}
\label{experiments}
\subsection{Experimental Settings}
\textbf{Datasets.}  
The REDS dataset \cite{nah2019ntire} is employed for training. This dataset contains 720p video sequences with diverse scenes and large inter-frame motion, which makes it particularly suitable for video super-resolution. For evaluation, the commonly used REDS4 dataset \cite{nah2019ntire} is adopted. High-resolution frames are downsampled by $\times4$ to form LR inputs and then encoded with H.264 codec ~\cite{wiegand2003overview} in capped constant rate factor (CRF) mode (CRF 18/23/28) using FFmpeg. This setting limits excessive bitrate and better reflects online video conditions. During decoding, we parse the bitstream to extract motion vectors, residual maps, and frame types as auxiliary information. To further evaluate the robustness of the proposed CDA-VSR across different resolutions, four sequences are randomly selected from the Inter4K dataset \cite{stergiou2022adapool}, each containing 300 frames with resolutions of 2K, 1080p, and 720p. Following the same processing and compression settings as REDS4, the sequences are encoded using Capped-CRF with CRF=23. 

\noindent \textbf{Implementation Details.}
We adopt 3 residual blocks (RBs) \cite{lim2017enhanced} for feature extraction. For reconstruction, 24 RBs are employed for I-frames, while 12 RBs are used for P-frames. The number of channels in the convolutional layer is 64. The model is trained using the Adam \cite{kinga2015method} optimizer with $\beta_1 = 0.9$ and $\beta_2 = 0.99$. The learning rate is initialized to $2\times10^{-4}$ and gradually decayed with the Cosine Annealing scheme \cite{loshchilov2016sgdr}. We train on 15-frame clips with $64{\times}64$ random crops and random horizontal flips/rotations, for 300k iterations with batch size 8. The model is implemented in PyTorch and trained on an NVIDIA RTX 3090.

\subsection{Experiment on the REDS4 Dataset}

We evaluate the proposed CDA-VSR against several open-source state-of-the-art online VSR methods, including RRN~\cite{isobe2020revisiting}, RSDN~\cite{isobe2020video}, SSL-uni~\cite{xia2023structured}, KSNet-uni~\cite{jin2023kernel}, MMVSR~\cite{tang2025multi} and TMP~\cite{zhang2024tmp}. Additionally, we construct a variant \emph{BasicVSR$^{*}$} by eliminating the backward propagation branch of BasicVSR~\cite{chan2021basicvsr} in order to satisfy the causality and latency requirements of online VSR applications. All methods are trained and evaluated under the same experimental settings as our method to ensure a fair comparison. 

\noindent \textbf{Quantitative results.} Table~\ref{tab1} shows the quantitative comparison results under three compression levels (CRF18/23/28). As shown, our CDA-VSR achieves the best PSNR/SSIM across all compression levels while attaining the lowest runtime and MACs. At CRF28, our CDA-VSR outperforms TMP by +0.13 dB and exceeds its speed by over 2$\times$. The number of parameters in CDA-VSR is higher because the reconstruction module has two branches. During inference, only one of these branches is activated for each frame depending on its type, so the effective runtime and computational cost remain minimal.

\noindent \textbf{Qualitative results.} We conduct visual comparisons with several representative methods on the REDS4 dataset, as shown in Figure~\ref{fig4}. Implicit alignment methods (e.g., RRN and RSDN) tend to produce blurry edges and lose fine structures. BasicVSR$^{*}$, SSL-uni, and TMP preserve sharper boundaries than implicit alignment methods, but still fail to recover fine textures. In contrast, our CDA-VSR reconstructs clearer edges and more detailed textures.

\begin{table}[t]
\centering
\caption{Comparison with state-of-the-art online VSR methods on the Inter4K dataset at different resolutions. The best and the second best results are colored with \textcolor{red}{red} and \textcolor{blue}{blue}.}
\label{tab2}
\resizebox{\linewidth}{!}{
\begin{tabular}{L{2.6cm}C{0.8cm}C{0.8cm}C{0.8cm} C{0.8cm}C{0.8cm}C{0.8cm} C{0.6cm}C{0.6cm}C{0.6cm}}
\toprule
\multirow{2}{*}{Method} & \multicolumn{3}{c}{PSNR (dB)$\uparrow$} & \multicolumn{3}{c}{SSIM$\uparrow$} & \multicolumn{3}{c}{FPS (1/s)$\uparrow$} \\
\cmidrule(lr){2-4} \cmidrule(lr){5-7} \cmidrule(lr){8-10}
 & 720p & 1080p & 2K & 720p & 1080p & 2K & 720p & 1080p & 2K \\
\midrule
BasicVSR$^{*}$~\cite{chan2021basicvsr} & 26.88 & 28.36 & 29.73 & 0.7941 & 0.8454 & 0.8816 & 28.6 & 13.6 & 7.7 \\
RRN~\cite{isobe2020revisiting} & 26.61 & 28.07 & 29.33 & 0.7821 & 0.8353 & 0.8707 & \textcolor{blue}{58.3} & 26.5 & 15.2 \\
RSDN~\cite{isobe2020video} & 26.65 & 28.07 & 29.29 & 0.7839 & 0.8361 & 0.8709 & 26.2 & 11.9 & 6.7 \\
SSL-uni~\cite{xia2023structured} & 26.83 & 28.15 & 29.58 & 0.7916 & 0.8429 & 0.8783 & 48.1 & \textcolor{blue}{27.1} & \textcolor{blue}{16.9} \\
KSNet-uni \cite{jin2023kernel} & 26.85 & 28.28 & 29.61 & 0.7905 & 0.8405 & 0.8768 & 34.0 & 17.8 & 10.6 \\
MMVSR \cite{tang2025multi} & 26.72 & 28.20 & 29.51 & 0.7852 & 0.8388 & 0.8748 & 43.1 & 21.1 & 12.4 \\
TMP~\cite{zhang2024tmp} & \textcolor{blue}{26.95} & \textcolor{blue}{28.45} & \textcolor{blue}{29.76} & \textcolor{blue}{0.7959} & \textcolor{blue}{0.8473} & \textcolor{blue}{0.8822} & 45.7 & 20.8 & 11.4 \\
CDA-VSR (Ours) & \textcolor{red}{27.13} & \textcolor{red}{28.64} & \textcolor{red}{29.98} & \textcolor{red}{0.8022} & \textcolor{red}{0.8525} & \textcolor{red}{0.8868} & \textcolor{red}{92.6} & \textcolor{red}{44.2} & \textcolor{red}{25.1} \\
\bottomrule
\end{tabular}
}
\end{table}

\subsection{Experiments on the Inter4K Dataset at Different Resolutions}

To further evaluate the robustness of our method under higher resolutions, we conduct experiments on the Inter4K dataset, which contains sequences at 720p, 1080p, and 2K resolutions. Table~\ref{tab2} reports PSNR, SSIM, and FPS. Our method consistently achieves the best PSNR and SSIM at all resolutions. At 2K, our CDA-VSR achieves 29.98 dB, outperforming TMP by +0.22 dB. In terms of efficiency, the advantage grows with resolution: at 1080p we maintain film real-time with a clear margin, and at 2K we still exceed the 24 FPS threshold while all other methods fall well below it. Overall, these results confirm that CDA-VSR achieves a superior trade-off between reconstruction quality and runtime efficiency.

\subsection{Ablation Studies}
In this section, we conduct ablation experiments on three key components of CDA-VSR: the MV-guided deformable alignment (MVGDA), the residual map gated fusion (RMGF), and the frame-type-aware reconstruction (FTAR). All models are trained on the REDS dataset and evaluated on REDS4.

\begin{figure*}[t!]
	\centering
	\includegraphics[width=6.4 in]{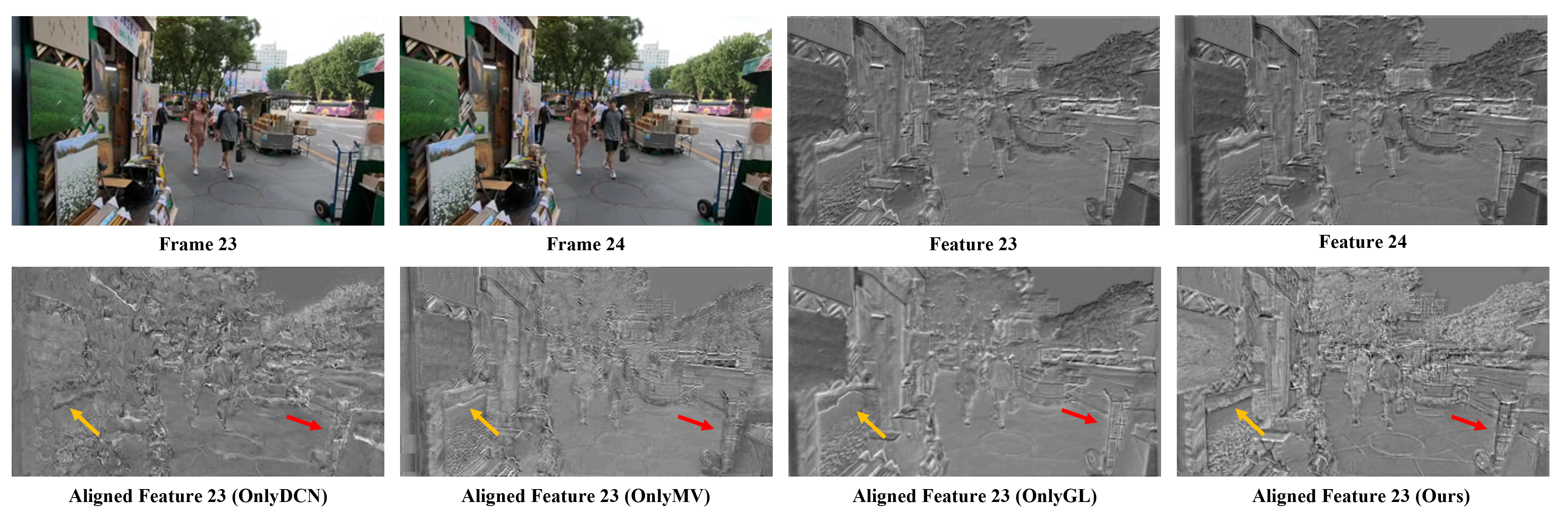}
	\caption{Feature map visualization of different alignment methods. Three variants (\emph{OnlyMV}, \emph{OnlyDCN}, \emph{OnlyGL}) and our MV-guided deformable alignment (MVGDA) are compared by visualizing the feature maps before and after alignment. For clarity, only the first channel of each feature map is shown.}
	\label{fig6}
\end{figure*}

\begin{table}[t]
\centering
\caption{Ablation study on the MV-guided deformable alignment (MVGDA) using the REDS4 dataset. 
The best results are colored with \textcolor{red}{red}.}
\label{tab:ablation_alignment}
\resizebox{\linewidth}{!}{
\begin{tabular}{lcccccc}
\toprule
\multirow{2}{*}{Method} & \multirow{2}{*}{\shortstack{Runtime$\downarrow$ \\ (ms)}} & \multirow{2}{*}{\shortstack{Params$\downarrow$ \\ (M)}} & \multicolumn{3}{c}{PSNR(dB)$\uparrow$ / SSIM$\uparrow$} \\
\cmidrule(lr){4-6}
& & & CRF18 & CRF23 & CRF28 \\
\midrule
OnlyMV    & \textcolor{red}{10.2}  & \textcolor{red}{3.17} & 27.59/0.7724 & 26.56/0.7332 & 25.19/0.6825 \\
OnlyDCN   & 10.6  & 3.28 & 27.35/0.7638 & 26.38/0.7263 & 25.06/0.6769 \\
OnlyGL    & 15.5  & 4.61 & 27.73/0.7761 & 26.66/0.7364 & 25.25/0.6844 \\
MVGDA (Ours) & 10.8  & 3.28 & \textcolor{red}{27.76}/\textcolor{red}{0.7779} & \textcolor{red}{26.70}/\textcolor{red}{0.7384} & \textcolor{red}{25.30}/\textcolor{red}{0.6869} \\
\bottomrule
\end{tabular}
}
\end{table}
\noindent \textbf{Effectiveness of MVGDA.}
To validate the effectiveness of the MV-guided deformable alignment (MVGDA), 
we design three variants: \emph{OnlyMV} with motion-vector warping only, 
\emph{OnlyDCN} with deformable convolution only, and \emph{OnlyGL} with optical flow warping only. 
All are retrained under the same settings, and the results are given in Table~\ref{tab:ablation_alignment}. 
\emph{OnlyMV} is fastest but drops 0.17 dB at CRF18.  
\emph{OnlyDCN} drops 0.41 dB at CRF18.  
\emph{OnlyGL} attains competitive quality, but requires about 1.4$\times$ the runtime of our method.  
MVGDA achieves the best quality with low runtime, striking a favorable balance between accuracy and efficiency.
To further explain these results, we visualize the feature maps before and after alignment, as illustrated in Figure~\ref{fig6}.
\emph{OnlyDCN} struggles with large motions, leading to blurred and misaligned structures in dynamic regions. 
\emph{OnlyMV} provides efficient global motion alignment but produces block-level discontinuities at object boundaries, as indicated by the red arrows.
\emph{OnlyGL} provides accurate motion compensation and preserves fine structures; however, residual misalignments persist in the regions indicated by the green arrows.
In contrast, our MVGDA produces cleaner and more consistent features across frames, effectively reducing misalignment errors. 

\begin{table}[t]
\centering
\caption{Ablation study on the residual map gated fusion (RMGF) using the REDS4 dataset. 
The best results are colored with \textcolor{red}{red}.}
\label{tab:ablation_fusion}
\resizebox{\linewidth}{!}{
\begin{tabular}{lcccccc}
\toprule
\multirow{2}{*}{Method} & \multirow{2}{*}{\shortstack{Runtime$\downarrow$ \\ (ms)}} & \multirow{2}{*}{\shortstack{Params$\downarrow$ \\ (M)}} & \multicolumn{3}{c}{PSNR(dB)$\uparrow$ / SSIM$\uparrow$} \\
\cmidrule(lr){4-6}
& & & CRF18 & CRF23 & CRF28 \\
\midrule
NoGate    & \textcolor{red}{10.8}  & \textcolor{red}{3.26} & 27.63/0.7739 & 26.60/0.7347 & 25.22/0.6838 \\
RMGF (Ours)  & \textcolor{red}{10.8}  & 3.28 & \textcolor{red}{27.76}/\textcolor{red}{0.7779} & \textcolor{red}{26.70}/\textcolor{red}{0.7384} & \textcolor{red}{25.30}/\textcolor{red}{0.6869} \\
\bottomrule
\end{tabular}
}
\end{table}

\begin{figure}[t!]
	\centering
	\includegraphics[width=3.2 in]{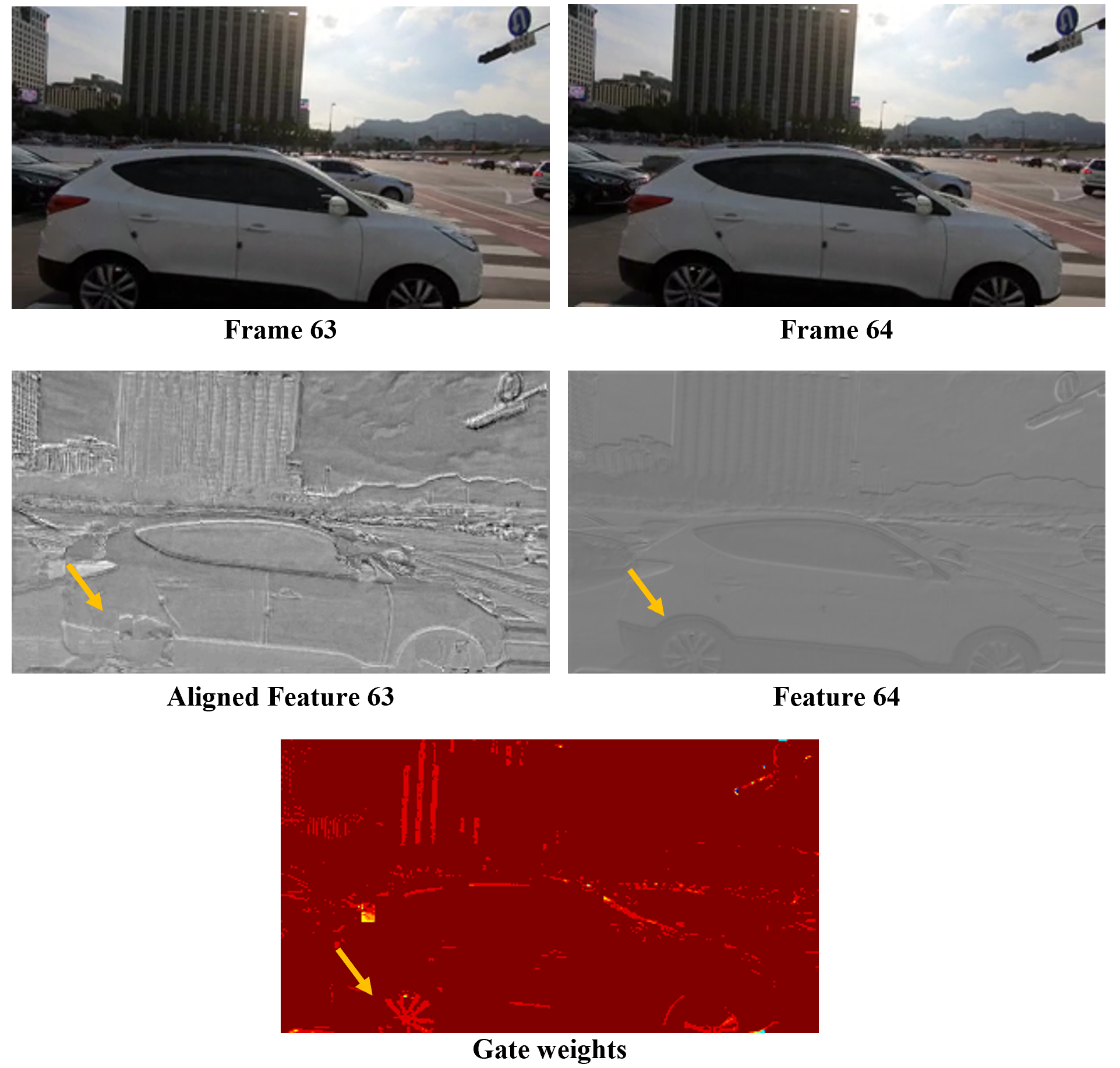}
	\caption{Heatmap visualization of gated fusion weights. Warmer colors (red) indicate larger gating values. The gate performs spatially selective fusion by assigning strong responses to stable regions while suppressing misaligned areas (green arrows).}
	\label{fig7}
\end{figure}

\noindent \textbf{Effectiveness of RMGF.}
To assess the residual map gated fusion (RMGF), we build a no-gating variant (named \emph{NoGate}). As shown in Table~\ref{tab:ablation_fusion}, RMGF outperforms NoGate at all compression levels. At CRF18, NoGate attains 27.63 dB, which is 0.13 dB lower than our full model. The gain comes from gating that regulates temporal fusion: simple concatenation propagates misaligned details, whereas RMGF upweights consistent regions and suppresses unreliable ones. To further corroborate this, we visualize the gate weights as heatmaps. As shown in Figure~\ref{fig7}, the gate assigns higher weights to previous-frame features in well-aligned, temporally stable regions (e.g., the car body) while suppressing misaligned areas such as the rotating wheels indicated by the green arrows. These results confirm that residual maps provide a valuable cue for selective temporal fusion, improving reconstruction quality.

\begin{table}[t]
\centering
\caption{Ablation study on the frame-type-aware reconstruction (FTAR) using the REDS4 dataset. }
\label{tab:ablation_reconstruction}
\resizebox{\linewidth}{!}{
\begin{tabular}{lccccc}
\toprule
\multirow{2}{*}{Method} & \multirow{2}{*}{\shortstack{Runtime$\downarrow$ \\ (ms)}} & \multicolumn{3}{c}{PSNR(dB)$\uparrow$ / SSIM$\uparrow$} \\
\cmidrule(lr){3-5}
& &  CRF18 & CRF23 & CRF28 \\
\midrule
I=12,P=12    & 10.7   & 27.60/0.7724 & 26.57/0.7334 & 25.19/0.6824 \\
I=24,P=24    & 16.8  & 27.80/0.7786 & 26.74/0.7390 & 25.34/0.6875 \\
I=24,P=12 (Ours) & 10.8  & 27.76/0.7779 & 26.70/0.7384 &25.30/0.6869 \\
\bottomrule
\end{tabular}
}
\end{table}

\noindent \textbf{Effectiveness of FTAR.}
We evaluate frame-type-aware reconstruction (FTAR) on the REDS4 dataset. As shown in Table~\ref{tab:ablation_reconstruction}, deepening both branches (I=24, P=24) improves PSNR and SSIM over the shallow baseline (I=12, P=12) at all CRFs, but increases latency by 57\% (10.7 ms to 16.8 ms). In contrast, the FTAR setting (I=24, P=12) preserves most of the accuracy gains with almost no extra cost: relative to (I=12, P=12), it adds +0.16/+0.13/+0.11 dB PSNR and +0.0055/+0.0050/+0.0045 SSIM, with only +0.1 ms overhead. The consistent trend across CRFs indicates that allocating more capacity to I-frames captures most of the quality benefit, while keeping the P-frame branch lightweight avoids redundant computation and yields a trade-off between accuracy and efficiency for online VSR.

\section{Conclusion}
In this paper, we propose a compressed-domain-aware framework for online VSR, termed CDA-VSR. Unlike existing methods that operate only on LR frames, CDA-VSR leverages compressed-domain information through three dedicated modules. We develop a motion-vector-guided deformable alignment module that uses motion vectors for coarse warping and learns only local residual offsets, which greatly reduces the cost of offset estimation while maintaining accuracy. We further present a residual map gated fusion module that predicts spatial weights to suppress mismatched regions and emphasize reliable details. We also introduce a frame-type-aware reconstruction scheme that allocates higher capacity to I-frames while keeping a lightweight branch for P-frames to better balance accuracy and efficiency. Extensive experiments demonstrate that CDA-VSR achieves higher super-resolution quality and faster inference speed than current state-of-the-art online VSR methods, delivering more than $2\times$ the FPS. In future work, we plan to extend the use of compressed-domain information to broader video restoration and enhancement tasks, such as artifact removal and temporal interpolation.
{
    \small
    \bibliographystyle{ieeenat_fullname}
    \bibliography{main}

@String(BMVC= {Brit. Mach. Vis. Conf.})

@String(ICLR = {Int. Conf. Learn. Represent.})

@String(AAAI = {AAAI})

@String(CVPRW= {IEEE Conf. Comput. Vis. Pattern Recog. Worksh.})

@String(BMVC  =	{BMVC})

@String(ICLR  = {ICLR})

@String(CVPRW= {CVPRW})

@inproceedings{fuoli2023fast,
  title={Fast online video super-resolution with deformable attention pyramid},
  author={Fuoli, Dario and Danelljan, Martin and Timofte, Radu and Van Gool, Luc},
  booktitle={Proceedings of the IEEE/CVF winter conference on applications of computer vision},
  pages={1735--1744},
  year={2023}
}

@article{xiao2023online,
  title={Online video super-resolution with convolutional kernel bypass grafts},
  author={Xiao, Jun and Jiang, Xinyang and Zheng, Ningxin and Yang, Huan and Yang, Yifan and Yang, Yuqing and Li, Dongsheng and Lam, Kin-Man},
  journal={IEEE Transactions on Multimedia},
  volume={25},
  pages={8972--8987},
  year={2023},
  publisher={IEEE}
}

@article{zhang2024tmp,
  title={Tmp: Temporal motion propagation for online video super-resolution},
  author={Zhang, Zhengqiang and Li, Ruihuang and Guo, Shi and Cao, Yang and Zhang, Lei},
  journal={IEEE Transactions on Image Processing},
  year={2024},
  publisher={IEEE}
}

@article{yin2024online,
  title={Online streaming video super-resolution with convolutional look-up table},
  author={Yin, Guanghao and Qu, Zefan and Jiang, Xinyang and Jiang, Shan and Han, Zhenhua and Zheng, Ningxin and Yang, Huan and Liu, Xiaohong and Yang, Yuqing and Li, Dongsheng and others},
  journal={IEEE Transactions on Image Processing},
  volume={33},
  pages={2305--2317},
  year={2024},
  publisher={IEEE}
}

@inproceedings{sajjadi2018frame,
  title={Frame-recurrent video super-resolution},
  author={Sajjadi, Mehdi SM and Vemulapalli, Raviteja and Brown, Matthew},
  booktitle={Proceedings of the IEEE conference on computer vision and pattern recognition},
  pages={6626--6634},
  year={2018}
}

@inproceedings{isobe2020video,
  title={Video super-resolution with recurrent structure-detail network},
  author={Isobe, Takashi and Jia, Xu and Gu, Shuhang and Li, Songjiang and Wang, Shengjin and Tian, Qi},
  booktitle={European conference on computer vision},
  pages={645--660},
  year={2020},
  organization={Springer}
}

@inproceedings{isobe2020revisiting,
  title={Revisiting Temporal Modeling for Video Super-resolution},
  author={Isobe, Takashi and Zhu, Fang and Wang, Shengjin},
  booktitle={BMVC},
  year={2020}
}

@inproceedings{fuoli2019efficient,
  title={Efficient video super-resolution through recurrent latent space propagation},
  author={Fuoli, Dario and Gu, Shuhang and Timofte, Radu},
  booktitle={2019 IEEE/CVF International Conference on Computer Vision Workshop (ICCVW)},
  pages={3476--3485},
  year={2019},
  organization={IEEE}
}

@inproceedings{chen2021fast,
  title={Fast object detection in hevc intra compressed domain},
  author={Chen, Liuhong and Sun, Heming and Katto, Jiro and Zeng, Xiaoyang and Fan, Yibo},
  booktitle={2021 29th European Signal Processing Conference (EUSIPCO)},
  pages={756--760},
  year={2021},
  organization={IEEE}
}

@article{liu2024vadiffusion,
  title={Vadiffusion: Compressed domain information guided conditional diffusion for video anomaly detection},
  author={Liu, Hao and He, Lijun and Zhang, Miao and Li, Fan},
  journal={IEEE Transactions on Circuits and Systems for Video Technology},
  volume={34},
  number={9},
  pages={8398--8411},
  year={2024},
  publisher={IEEE}
}

@article{chen2021compressed,
  title={Compressed domain deep video super-resolution},
  author={Chen, Peilin and Yang, Wenhan and Wang, Meng and Sun, Long and Hu, Kangkang and Wang, Shiqi},
  journal={IEEE Transactions on Image Processing},
  volume={30},
  pages={7156--7169},
  year={2021},
  publisher={IEEE}
}

@inproceedings{zhang2022codec,
  title={A codec information assisted framework for efficient compressed video super-resolution},
  author={Zhang, Hengsheng and Zou, Xueyi and Guo, Jiaming and Yan, Youliang and Xie, Rong and Song, Li},
  booktitle={European Conference on Computer Vision},
  pages={220--235},
  year={2022},
  organization={Springer}
}

@inproceedings{wang2023compression,
  title={Compression-aware video super-resolution},
  author={Wang, Yingwei and Isobe, Takashi and Jia, Xu and Tao, Xin and Lu, Huchuan and Tai, Yu-Wing},
  booktitle={Proceedings of the IEEE/CVF Conference on Computer Vision and Pattern Recognition},
  pages={2012--2021},
  year={2023}
}

@inproceedings{xia2023structured,
  title={Structured sparsity learning for efficient video super-resolution},
  author={Xia, Bin and He, Jingwen and Zhang, Yulun and Wang, Yitong and Tian, Yapeng and Yang, Wenming and Van Gool, Luc},
  booktitle={Proceedings of the IEEE/CVF conference on computer vision and pattern recognition},
  pages={22638--22647},
  year={2023}
}

@inproceedings{chan2021basicvsr,
  title={Basicvsr: The search for essential components in video super-resolution and beyond},
  author={Chan, Kelvin CK and Wang, Xintao and Yu, Ke and Dong, Chao and Loy, Chen Change},
  booktitle={Proceedings of the IEEE/CVF conference on computer vision and pattern recognition},
  pages={4947--4956},
  year={2021}
}

@article{wu2024real,
  title={Real-time lightweight video super-resolution with RRED-based perceptual constraint},
  author={Wu, Xinyi and L{\'o}pez-Tapia, Santiago and Wang, Xijun and Molina, Rafael and Katsaggelos, Aggelos K},
  journal={IEEE Transactions on Circuits and Systems for Video Technology},
  volume={34},
  number={10},
  pages={10310--10325},
  year={2024},
  publisher={IEEE}
}

@inproceedings{viswanathan2025low,
  title={Low-Resource Video Super-Resolution using Memory, Wavelets, and Deformable Convolutions},
  author={Viswanathan, Kavitha and Sethi, Amit and Pathak, Shashwat and Bharambe, Piyush and Choudhary, Harsh},
  booktitle={Proceedings of the Computer Vision and Pattern Recognition Conference},
  pages={3444--3453},
  year={2025}
}

@article{wiegand2003overview,
  title={Overview of the H. 264/AVC video coding standard},
  author={Wiegand, Thomas and Sullivan, Gary J and Bjontegaard, Gisle and Luthra, Ajay},
  journal={IEEE Transactions on circuits and systems for video technology},
  volume={13},
  number={7},
  pages={560--576},
  year={2003},
  publisher={IEEE}
}

@inproceedings{tian2020tdan,
  title={Tdan: Temporally-deformable alignment network for video super-resolution},
  author={Tian, Yapeng and Zhang, Yulun and Fu, Yun and Xu, Chenliang},
  booktitle={Proceedings of the IEEE/CVF conference on computer vision and pattern recognition},
  pages={3360--3369},
  year={2020}
}

@inproceedings{wang2019edvr,
  title={EDVR: Video Restoration With Enhanced Deformable Convolutional Networks},
  author={Wang, Xintao and Chan, Kelvin CK and Yu, Ke and Dong, Chao and Loy, Chen Change},
  booktitle={2019 IEEE/CVF Conference on Computer Vision and Pattern Recognition Workshops (CVPRW)},
  pages={1954--1963},
  year={2019},
  organization={IEEE}
}

@inproceedings{li2020mucan,
  title={Mucan: Multi-correspondence aggregation network for video super-resolution},
  author={Li, Wenbo and Tao, Xin and Guo, Taian and Qi, Lu and Lu, Jiangbo and Jia, Jiaya},
  booktitle={European conference on computer vision},
  pages={335--351},
  year={2020},
  organization={Springer}
}

@inproceedings{caballero2017real,
  title={Real-time video super-resolution with spatio-temporal networks and motion compensation},
  author={Caballero, Jose and Ledig, Christian and Aitken, Andrew and Acosta, Alejandro and Totz, Johannes and Wang, Zehan and Shi, Wenzhe},
  booktitle={Proceedings of the IEEE conference on computer vision and pattern recognition},
  pages={4778--4787},
  year={2017}
}

@incollection{kappeler2016video,
  title={Video Super-Resolution With Convolutional Neural Networks},
  author={Kappeler, Armin and Yoo, Seunghwan and Dai, Qiqin and Katsaggelos, Aggelos K},
  booktitle={IEEE Transactions on Computational Imaging},
  pages={109--122},
  year={2016},
  publisher={IEEE}
}

@article{xue2019video,
  title={Video enhancement with task-oriented flow},
  author={Xue, Tianfan and Chen, Baian and Wu, Jiajun and Wei, Donglai and Freeman, William T},
  journal={International Journal of Computer Vision},
  volume={127},
  number={8},
  pages={1106--1125},
  year={2019},
  publisher={Springer}
}

@article{wang2020deep,
  title={Deep video super-resolution using HR optical flow estimation},
  author={Wang, Longguang and Guo, Yulan and Liu, Li and Lin, Zaiping and Deng, Xinpu and An, Wei},
  journal={IEEE Transactions on Image Processing},
  volume={29},
  pages={4323--4336},
  year={2020},
  publisher={IEEE}
}

@inproceedings{yi2019progressive,
  title={Progressive fusion video super-resolution network via exploiting non-local spatio-temporal correlations},
  author={Yi, Peng and Wang, Zhongyuan and Jiang, Kui and Jiang, Junjun and Ma, Jiayi},
  booktitle={Proceedings of the IEEE/CVF international conference on computer vision},
  pages={3106--3115},
  year={2019}
}

@inproceedings{li2024savsr,
  title={SAVSR: arbitrary-scale video super-resolution via a learned scale-adaptive network},
  author={Li, Zekun and Liu, Hongying and Shang, Fanhua and Liu, Yuanyuan and Wan, Liang and Feng, Wei},
  booktitle={Proceedings of the AAAI Conference on Artificial Intelligence},
  volume={38},
  number={4},
  pages={3288--3296},
  year={2024}
}

@inproceedings{chan2022basicvsr++,
  title={Basicvsr++: Improving video super-resolution with enhanced propagation and alignment},
  author={Chan, Kelvin CK and Zhou, Shangchen and Xu, Xiangyu and Loy, Chen Change},
  booktitle={Proceedings of the IEEE/CVF conference on computer vision and pattern recognition},
  pages={5972--5981},
  year={2022}
}

@article{dong2015image,
  title={Image super-resolution using deep convolutional networks},
  author={Dong, Chao and Loy, Chen Change and He, Kaiming and Tang, Xiaoou},
  journal={IEEE transactions on pattern analysis and machine intelligence},
  volume={38},
  number={2},
  pages={295--307},
  year={2015},
  publisher={IEEE}
}

@inproceedings{wang2023omni,
  title={Omni aggregation networks for lightweight image super-resolution},
  author={Wang, Hang and Chen, Xuanhong and Ni, Bingbing and Liu, Yutian and Liu, Jinfan},
  booktitle={Proceedings of the IEEE/CVF Conference on Computer Vision and Pattern Recognition},
  pages={22378--22387},
  year={2023}
}

@inproceedings{zheng2024smfanet,
  title={SMFANet: A lightweight self-modulation feature aggregation network for efficient image super-resolution},
  author={Zheng, Mingjun and Sun, Long and Dong, Jiangxin and Pan, Jinshan},
  booktitle={European conference on computer vision},
  pages={359--375},
  year={2024},
  organization={Springer}
}

@inproceedings{liang2021swinir,
  title={Swinir: Image restoration using swin transformer},
  author={Liang, Jingyun and Cao, Jiezhang and Sun, Guolei and Zhang, Kai and Van Gool, Luc and Timofte, Radu},
  booktitle={Proceedings of the IEEE/CVF international conference on computer vision},
  pages={1833--1844},
  year={2021}
}

@article{dong2025lightweight,
  title={Lightweight Real-World Image Super-Resolution via Channel Redundancy for Edge IoT Devices},
  author={Dong, Zhetao and Hou, Shujuan and Li, Hai and Wang, Yuhang and Gao, Ruixue},
  journal={IEEE Internet of Things Journal},
  year={2025},
  publisher={IEEE}
}

@article{shi2022rethinking,
  title={Rethinking alignment in video super-resolution transformers},
  author={Shi, Shuwei and Gu, Jinjin and Xie, Liangbin and Wang, Xintao and Yang, Yujiu and Dong, Chao},
  journal={Advances in Neural Information Processing Systems},
  volume={35},
  pages={36081--36093},
  year={2022}
}

@inproceedings{zhou2024video,
  title={Video super-resolution transformer with masked inter\&intra-frame attention},
  author={Zhou, Xingyu and Zhang, Leheng and Zhao, Xiaorui and Wang, Keze and Li, Leida and Gu, Shuhang},
  booktitle={Proceedings of the IEEE/CVF Conference on Computer Vision and Pattern Recognition},
  pages={25399--25408},
  year={2024}
}

@inproceedings{zhang2024realviformer,
  title={Realviformer: Investigating attention for real-world video super-resolution},
  author={Zhang, Yuehan and Yao, Angela},
  booktitle={European Conference on Computer Vision},
  pages={412--428},
  year={2024},
  organization={Springer}
}

@inproceedings{zhou2024upscale,
  title={Upscale-a-video: Temporal-consistent diffusion model for real-world video super-resolution},
  author={Zhou, Shangchen and Yang, Peiqing and Wang, Jianyi and Luo, Yihang and Loy, Chen Change},
  booktitle={Proceedings of the IEEE/CVF Conference on Computer Vision and Pattern Recognition},
  pages={2535--2545},
  year={2024}
}

@inproceedings{nah2019ntire,
  title={NTIRE 2019 Challenge on Video Deblurring and Super-Resolution: Dataset and Study},
  author={Nah, Seungjun and Baik, Sungyong and Hong, Seokil and Moon, Gyeongsik and Son, Sanghyun and Timofte, Radu and Lee, Kyoung Mu},
  booktitle={2019 IEEE/CVF Conference on Computer Vision and Pattern Recognition Workshops (CVPRW)},
  pages={1996--2005},
  year={2019},
  organization={IEEE}
}

@article{stergiou2022adapool,
  title={Adapool: Exponential adaptive pooling for information-retaining downsampling},
  author={Stergiou, Alexandros and Poppe, Ronald},
  journal={IEEE Transactions on Image Processing},
  volume={32},
  pages={251--266},
  year={2022},
  publisher={IEEE}
}

@inproceedings{jin2023kernel,
  title={Kernel dimension matters: To activate available kernels for real-time video super-resolution},
  author={Jin, Shuo and Liu, Meiqin and Yao, Chao and Lin, Chunyu and Zhao, Yao},
  booktitle={Proceedings of the 31st ACM International Conference on Multimedia},
  pages={8617--8625},
  year={2023}
}

@article{tang2025multi,
  title={Multi-Memory Streams: A Paradigm for Online Video Super-Resolution in Complex Exposure Scenes},
  author={Tang, Guozhi and Ge, Hongwei and Luo, Yong and Li, Bo and Wu, Chunguo},
  journal={IEEE Transactions on Multimedia},
  year={2025},
  publisher={IEEE}
}

@article{menon2024video,
  title={Video Super-Resolution for Optimized Bitrate and Green Online Streaming},
  author={Menon, Vignesh V and Rajendran, Prajit T and Premkumar, Amritha and Bross, Benjamin and Marpe, Detlev},
  journal={arXiv preprint arXiv:2402.03513},
  year={2024}
}

@article{zhu2025trajectory,
  title={Trajectory-aware Shifted State Space Models for Online Video Super-Resolution},
  author={Zhu, Qiang and Meng, Xiandong and Jiang, Yuxian and Zhang, Fan and Bull, David and Zhu, Shuyuan and Zeng, Bing},
  journal={arXiv preprint arXiv:2508.10453},
  year={2025}
}

@inproceedings{kinga2015method,
  title={A method for stochastic optimization},
  author={Kinga, Diederik and Adam, Jimmy Ba and others},
  booktitle={International conference on learning representations (ICLR)},
  volume={5},
  number={6},
  year={2015},
  organization={California;}
}

@inproceedings{lai2017deep,
  title={Deep laplacian pyramid networks for fast and accurate super-resolution},
  author={Lai, Wei-Sheng and Huang, Jia-Bin and Ahuja, Narendra and Yang, Ming-Hsuan},
  booktitle={Proceedings of the IEEE conference on computer vision and pattern recognition},
  pages={624--632},
  year={2017}
}

@article{loshchilov2016sgdr,
  title={Sgdr: Stochastic gradient descent with warm restarts},
  author={Loshchilov, Ilya and Hutter, Frank},
  journal={arXiv preprint arXiv:1608.03983},
  year={2016}
}

@inproceedings{xu2025videogigagan,
  title={Videogigagan: Towards detail-rich video super-resolution},
  author={Xu, Yiran and Park, Taesung and Zhang, Richard and Zhou, Yang and Shechtman, Eli and Liu, Feng and Huang, Jia-Bin and Liu, Difan},
  booktitle={Proceedings of the Computer Vision and Pattern Recognition Conference},
  pages={2139--2149},
  year={2025}
}

@inproceedings{shang2024arbitrary,
  title={Arbitrary-Scale Video Super-Resolution with Structural and Textural Priors},
  author={Shang, Wei and Ren, Dongwei and Zhang, Wanying and Fang, Yuming and Zuo, Wangmeng and Ma, Kede},
  booktitle={European Conference on Computer Vision},
  pages={73--90},
  year={2024},
  organization={Springer}
}

@article{xie2025star,
  title={Star: Spatial-temporal augmentation with text-to-video models for real-world video super-resolution},
  author={Xie, Rui and Liu, Yinhong and Zhou, Penghao and Zhao, Chen and Zhou, Jun and Zhang, Kai and Zhang, Zhenyu and Yang, Jian and Yang, Zhenheng and Tai, Ying},
  journal={arXiv preprint arXiv:2501.02976},
  year={2025}
}

@inproceedings{du2025patchvsr,
  title={PatchVSR: Breaking Video Diffusion Resolution Limits with Patch-wise Video Super-Resolution},
  author={Du, Shian and Xia, Menghan and Liu, Chang and Wang, Xintao and Wang, Jing and Wan, Pengfei and Zhang, Di and Ji, Xiangyang},
  booktitle={Proceedings of the Computer Vision and Pattern Recognition Conference},
  pages={17799--17809},
  year={2025}
}

@article{liang2024vrt,
  title={Vrt: A video restoration transformer},
  author={Liang, Jingyun and Cao, Jiezhang and Fan, Yuchen and Zhang, Kai and Ranjan, Rakesh and Li, Yawei and Timofte, Radu and Van Gool, Luc},
  journal={IEEE Transactions on Image Processing},
  volume={33},
  pages={2171--2182},
  year={2024},
  publisher={IEEE}
}

@inproceedings{shi2025self,
  title={Self-supervised ControlNet with Spatio-Temporal Mamba for Real-world Video Super-resolution},
  author={Shi, Shijun and Xu, Jing and Lu, Lijing and Li, Zhihang and Hu, Kai},
  booktitle={Proceedings of the Computer Vision and Pattern Recognition Conference},
  pages={7385--7395},
  year={2025}
}

@article{liang2022recurrent,
  title={Recurrent video restoration transformer with guided deformable attention},
  author={Liang, Jingyun and Fan, Yuchen and Xiang, Xiaoyu and Ranjan, Rakesh and Ilg, Eddy and Green, Simon and Cao, Jiezhang and Zhang, Kai and Timofte, Radu and Gool, Luc V},
  journal={Advances in Neural Information Processing Systems},
  volume={35},
  pages={378--393},
  year={2022}
}

@inproceedings{liu2025catanet,
  title={CATANet: Efficient Content-Aware Token Aggregation for Lightweight Image Super-Resolution},
  author={Liu, Xin and Liu, Jie and Tang, Jie and Wu, Gangshan},
  booktitle={Proceedings of the Computer Vision and Pattern Recognition Conference},
  pages={17902--17912},
  year={2025}
}

@inproceedings{wang2019fast,
  title={Fast object detection in compressed video},
  author={Wang, Shiyao and Lu, Hongchao and Deng, Zhidong},
  booktitle={Proceedings of the IEEE/CVF international conference on computer vision},
  pages={7104--7113},
  year={2019}
}

@inproceedings{fang2023you,
  title={You can ground earlier than see: An effective and efficient pipeline for temporal sentence grounding in compressed videos},
  author={Fang, Xiang and Liu, Daizong and Zhou, Pan and Nan, Guoshun},
  booktitle={Proceedings of the IEEE/CVF conference on computer vision and pattern recognition},
  pages={2448--2460},
  year={2023}
}

@inproceedings{lim2017enhanced,
  title={Enhanced deep residual networks for single image super-resolution},
  author={Lim, Bee and Son, Sanghyun and Kim, Heewon and Nah, Seungjun and Mu Lee, Kyoung},
  booktitle={Proceedings of the IEEE conference on computer vision and pattern recognition workshops},
  pages={136--144},
  year={2017}
}
}
\end{document}